\documentclass{article}
\usepackage{spconf,amsmath,graphicx,booktabs}
\usepackage{newtxtext,newtxmath}
\usepackage[final]{microtype}
\usepackage[table]{xcolor}
\usepackage[hidelinks]{hyperref}
\usepackage{url}
\usepackage{tikz}
\usepackage{pgfplots}
\usetikzlibrary{arrows.meta,positioning,fit,calc}
\usepgfplotslibrary{groupplots}
\pgfplotsset{compat=1.18}
\definecolor{nvblue}{HTML}{35618A}
\definecolor{nvorange}{HTML}{C9784C}
\definecolor{nvgreen}{HTML}{5F7B58}
\definecolor{nvpurple}{HTML}{766A8E}
\definecolor{nvgrid}{HTML}{D9D5CC}
\definecolor{nvink}{HTML}{353331}
\pgfplotsset{
  nvaxis/.style={
    axis line style={draw=nvink!70},
    tick style={draw=nvink!70},
    tick label style={font=\small,text=nvink},
    label style={font=\small,text=nvink},
    title style={font=\small\bfseries,text=nvink},
    grid=major,
    grid style={draw=nvgrid,line width=0.25pt},
    legend style={draw=none,fill=none,font=\small,/tikz/every even column/.append style={column sep=3pt}},
    every axis plot/.append style={line width=1.15pt,mark size=1.8pt},
  }
}

\title{A Systematic Study of Cross-Token Transfer for Data-Efficient Non-Verbal Text-to-Speech}
\name{Jing Pan \qquad Haibin Wu \qquad Minxue Niu \qquad Florian Metze \qquad Yuzong Liu}

\address{Meta, USA}

\begin{document}
\setcounter{topnumber}{4}
\setcounter{totalnumber}{6}
\renewcommand{\topfraction}{0.95}
\renewcommand{\textfraction}{0.05}
\clubpenalty=10000
\widowpenalty=10000
\displaywidowpenalty=10000
\maketitle

\begin{abstract}
Non-verbal vocalizations (NVs), such as laughter and breath, carry affective cues that words alone do not convey.
Yet NV data remains scarce because speech annotation pipelines often omit or normalize these events.
Existing work has improved NV text-to-speech through model design and data curation, but whether one NV token can help a model learn another has received little systematic study.
We investigate this \emph{cross-token transferability} in a controlled CosyVoice 2 setup.
We first trace a laughter-only scaling curve, then hold laughter at 200 or 1,600 examples while adding breath, sigh, cough, lip-smack, or neutral speech.
With only 200 laughter examples, adding 3,200 cough examples raises end-to-end laughter F1 from 0.170 to 0.529, surpassing the 0.495 obtained from 1,600 laughter examples alone.
At the 1,600-example anchor, cough reaches 0.695, nearly matching 6,400 laughter examples (0.697).
Waveform-level Whisper-D scoring and a semantic-token transcriber show consistent trends.
Cross-token transfer therefore offers a complementary route to low-resource NV generation.
\end{abstract}


\section{Introduction}

Spoken communication conveys more than words.
Non-verbal vocalizations (NVs) such as laughter, breaths, sighs, cries, and coughs reveal speaker state, affect, and conversational intent~\cite{provine2000laughter,scherer2003vocal}.
Laughter can signal warmth or empathy; filled pauses and breaths expose hesitation or control, while other bursts add rhythm and emphasis.
Words establish content, but NVs shape how an utterance is understood, making explicit control important for expressive and socially natural TTS~\cite{wu2024laugh}.
Unfortunately, the cues captured by a microphone often disappear during data preparation: conventional transcripts preserve lexical content while omitting NVs or collapsing them into inconsistent labels.
Rare events such as crying receive even fewer reliable annotations, leaving far less usable supervision than ordinary speech.

Recent work has expanded NV datasets and improved controllable generation, as reviewed in Section~\ref{sec:related_work}.
One complementary question remains open: can a well-resourced NV token improve a different, low-resource token during TTS training?
If cross-token transfer exists, abundant donor events could provide reusable acoustic structure, reduce the amount of target-specific annotation needed, and make rare NV capabilities cheaper to add.
Knowing which donors transfer would also help data curation prioritize useful examples rather than simply collecting more speech.

We study this question through controlled scaling and transfer experiments with laughter as the target event.
We first vary the amount of laughter data to establish its learning curve.
We then fix the laughter budget and add breath, sigh, cough, or lip-smack examples, with neutral speech as a control.
All conditions use the same CosyVoice 2 architecture and fine-tuning recipe, so observed differences can be attributed to training-data composition.

Our contributions are:
\begin{itemize}
    \item a controlled study of laughter scaling and cross-token transfer under a fixed TTS architecture;
    \item evidence that 3,200 cough examples raise laughter F1 from 0.170 to 0.529 when only 200 laughter examples are available, outperforming an eightfold increase in target-only data; and
    \item complementary evaluation with Whisper-D on generated waveforms and a learned transcriber that reads semantic tokens before waveform rendering.
\end{itemize}

\section{Related Work}
\label{sec:related_work}

\textbf{Data creation and NV-aware annotation.}
NV resources differ in how labels are obtained and how precisely events are aligned.
Early event-aware ASR decoded words and acoustic social signals jointly, establishing a unified transcript representation~\cite{inaguma2018joint,shione2023automatic}.
Recent pipelines turn recognition into a data-construction tool: NVSpeech bootstraps NV-aware annotation at scale, while WESR evaluates whether an event is both recognized and placed correctly within a transcript~\cite{liao2025nvspeech,yang2026wesr}.
Curated corpora offer finer supervision: NonverbalTTS aligns English NV and emotion labels, whereas SMIIP-NV and MNV-17 provide multi-level Mandarin annotations~\cite{borisov2025nonverbaltts,wu2025smiipnv,mai2026mnv17}.
NonVerbalSpeech-38K and SynParaSpeech trade some manual curation for scale through automatic annotation or generation~\cite{ye2025nonverbalspeech,bai2025synparaspeech}.
Beyond Words compares curriculum learning, data balancing, voice conversion, and auxiliary-token transfer for rare-event ASR~\cite{yang2026beyond}; NV-Bench and NVV-SuperBench expand synthesis evaluation across event types and quality dimensions~\cite{ni2026nvbench,xue2026nvvsuperbench}.

\textbf{Modeling.}
EmoCtrl-TTS introduces detector-derived conditioning and time-varying emotion embeddings to control laughter and crying within a flow-matching TTS model~\cite{wu2024laugh}.
Recent preference optimization instead uses an NV-aware recognition score to improve generation over 18 NV types~\cite{li2026preference}.
Both directions improve how a system represents or optimizes NV behavior; expanding and cleaning training corpora improves the evidence available to it.
Neither, however, isolates whether examples of one NV token help a scarce second token.
Our study is therefore complementary: we hold the architecture and optimization recipe fixed, vary only target--donor data composition, and measure the effect both before and after waveform rendering.

\begin{figure*}[t]
  \centering
  \begin{tikzpicture}[
  x=1.40cm,y=1cm,font=\small,
  >={Latex[length=1.8mm,width=1.2mm]},
  flow/.style={->,draw=nvink!75,line width=0.75pt,line cap=round,line join=round},
  contextflow/.style={->,draw=nvpurple!85,line width=0.75pt,dashed,line cap=round,line join=round},
  panel/.style={draw=nvgreen!75,line width=0.7pt,rounded corners=3pt,fill=white},
  box/.style={draw=nvink!60,line width=0.6pt,rounded corners=2pt,minimum height=7.5mm,align=center,inner xsep=3pt,inner ysep=2pt},
  text-box/.style={box,fill=nvgreen!10,text width=18mm},
  lm-box/.style={box,fill=nvorange!13,draw=nvorange,text width=25mm},
  token-box/.style={box,fill=nvblue!10,draw=nvblue,text width=21mm},
  decoder-box/.style={box,fill=nvpurple!10,draw=nvpurple,text width=28mm},
  prompt-box/.style={box,fill=nvgreen!9,draw=nvgreen,text width=23mm},
  eval-box/.style={box,fill=nvblue!7,draw=nvblue,text width=30mm,minimum height=8mm},
  metric-box/.style={box,fill=nvgreen!9,draw=nvgreen,text width=19mm,minimum height=13mm},
  study-box/.style={box,fill=nvorange!8,draw=nvorange,minimum height=7mm}
]

\draw[panel] (-0.28,0.12) rectangle (12.15,3.42);
\node[anchor=west,font=\small\bfseries,text=nvgreen] at (-0.08,3.16) {(a) CosyVoice generation path};
\node[text-box] (text) at (0.45,2.22) {Text tokens\\with NV markers};
\node[lm-box] (lm) at (2.75,2.22) {\textbf{Semantic LM}\\autoregressive text to speech tokens};
\node[token-box] (tokens) at (5.30,2.22) {Semantic tokens\\discrete speech content};
\node[decoder-box] (decoder) at (8.30,2.22) {Conditional flow + vocoder\\waveform rendering};
\node[box,fill=nvorange!8,draw=nvorange,text width=12mm] (audio) at (11.25,2.22) {Audio};
\node[prompt-box] (prompt) at (3.65,0.78) {Prompt speech\\speaker + context};
\node[token-box] (tokenizer) at (6.20,0.78) {Semantic tokenizer};

\draw[flow] (text) -- (lm);
\draw[flow] (lm) -- (tokens);
\draw[flow] (tokens) -- (decoder);
\draw[flow] (decoder) -- (audio);
\draw[contextflow] (prompt) -- (tokenizer);
\draw[contextflow] (tokenizer.north) -- ++(0,0.22) -| (lm.south);
\draw[contextflow] (prompt.south) -- ++(0,-0.08) -| (decoder.south);

\draw[panel] (-0.28,-2.50) rectangle (5.83,-0.10);
\node[anchor=west,font=\small\bfseries,text=nvorange] at (-0.08,-0.34) {(b) Cross-token experiments};
\node[study-box,text width=47mm] at (2.75,-0.82) {\textbf{Target scaling}: laughter, $0$--$6.4$k};
\node[study-box,text width=22mm] at (1.05,-1.86) {\textbf{Target}\\\texttt{ha-ha} laughter\\200 / 1.6k};
\node[font=\large\bfseries,text=nvink!75] at (2.77,-1.86) {$+$};
\node[study-box,text width=28mm] at (4.58,-1.86) {\textbf{Donor NV}\\cough, breath, sigh, lip-smack};

\draw[panel] (6.03,-2.50) rectangle (12.15,-0.10);
\node[anchor=west,font=\small\bfseries,text=nvblue] at (6.23,-0.34) {(c) Complementary evaluation};
\node[token-box,text width=18mm] (lmout) at (6.90,-1.35) {LM semantic tokens};
\node[eval-box] (tokeneval) at (9.10,-0.90) {Semantic-token eval.\\NV transcriber};
\node[eval-box] (waveeval) at (9.10,-1.78) {Waveform eval.\\vocoder + Whisper-D};
\node[metric-box] (metrics) at (11.38,-1.35) {NV transcript\\precision / recall / F1};
\draw[flow] (lmout.north east) -- (tokeneval.west);
\draw[flow] (lmout.south east) -- (waveeval.west);
\draw[flow] (tokeneval.east) -- (metrics.north west);
\draw[flow] (waveeval.east) -- (metrics.south west);
\end{tikzpicture}
  \caption{Study overview: (a) CosyVoice generation, (b) controlled target scaling and cross-token transfer, and (c) complementary waveform and semantic-token evaluation.}
  \label{fig:framework}
\end{figure*}
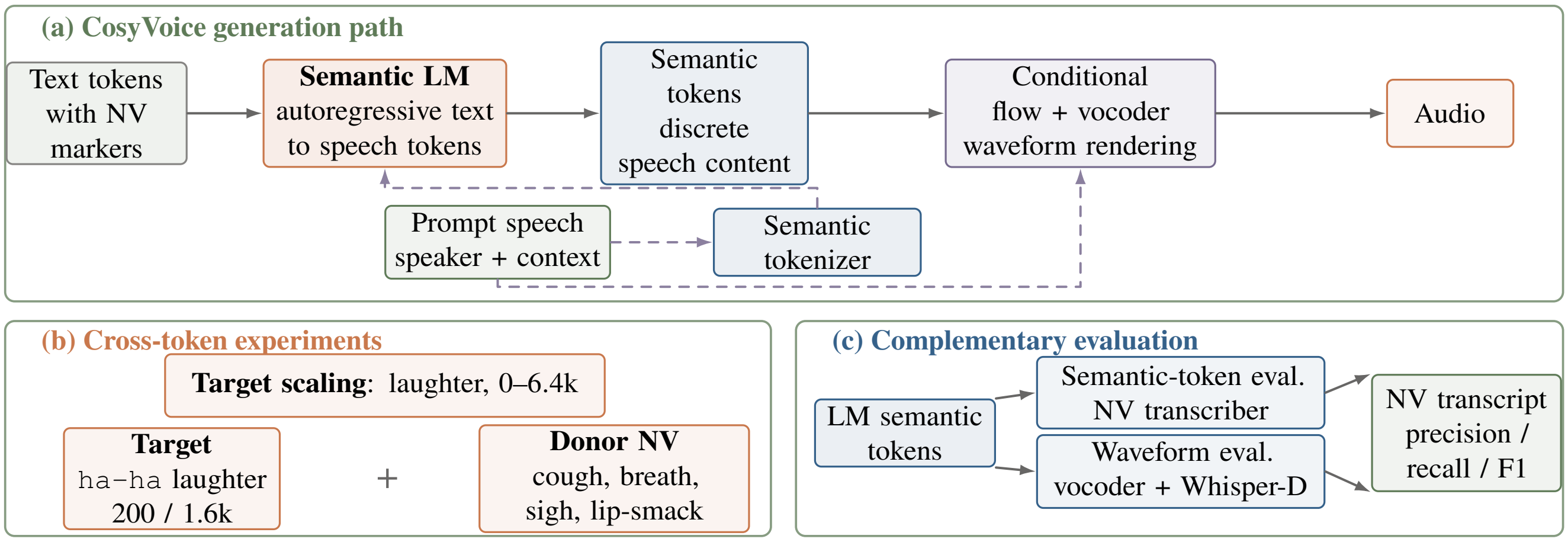
\section{Methodology}

\subsection{Framework}
We build on the CosyVoice 2 generation pipeline~\cite{du2024cosyvoice}, shown in Fig.~\ref{fig:framework}.
The input text contains lexical tokens and explicit NV markers such as \texttt{[laughter]}.
A prompt utterance supplies speaker and acoustic context.
CosyVoice 2 converts the prompt into discrete semantic tokens and conditions an autoregressive semantic language model (LM) on both the prompt tokens and input text.
The LM predicts the target semantic-token sequence\footnote{\small\href{https://github.com/xingchensong/S3Tokenizer}{github.com/xingchensong/S3Tokenizer}} $\boldsymbol{\mu}=(\mu_1,\ldots,\mu_L)$ with
\begin{equation}
\mathcal{L}_{\mathrm{sem}}=-\sum_{l=1}^{L}\log p(\mu_l\mid\mu_{<l},\mathbf{t}_{\mathrm{NV}},\mathbf{p}),
\end{equation}
where $\mathbf{t}_{\mathrm{NV}}$ is the text sequence with NV markers and $\mathbf{p}$ denotes the prompt representation.
A conditional flow decoder and vocoder render the predicted semantic tokens as a waveform.
In every experiment, we fine-tune only the semantic LM and reuse the same tokenizer, flow decoder, and vocoder.

\subsection{Case Studies}
Collecting a large corpus for every affective burst is costly.
Phonetic and physiological studies suggest that laughter, crying, and coughs draw on related respiratory and laryngeal mechanisms~\cite{provine2000laughter,scherer2003vocal}.
Laughter, for example, consists of rhythmic glottal pulses superimposed on forced exhalation~\cite{trouvain2014laughing}.
Although these events remain perceptually distinct, their shared airflow, voicing, and temporal structure creates a plausible basis for transfer.

Recent TTS evidence supports this possibility.
EmoCtrl-TTS showed that embeddings from a laughter detector can support fine-grained, controllable crying synthesis~\cite{wu2024laugh}.
This cross-event reuse motivates our data-level question: can examples labeled with an abundant donor token help the same semantic LM learn a scarce target token?

We test two controlled cases.
The \emph{target scaling} study trains on nested subsets containing 0, 200, 400, 800, 1,600, 3,200, or 6,400 laughter examples, establishing low-resource and near-saturation operating points.
The \emph{cross-token transfer} study fixes laughter at 200 or 1,600 examples and adds approximately 3,200 examples of breath, sigh, cough, or lip-smack.
A neutral-speech mixture tests whether arbitrary additional audio is sufficient.
At the 200-example anchor, cough ranges from 0 to 3,200 examples. All else remains fixed.

\subsection{Evaluation}
We measure whether the model generates laughter at the requested location using two complementary paths.

\textbf{Semantic-token transcriber.}
We train a separate autoregressive transcriber that maps the semantic-token sequence $\boldsymbol{\mu}$ to an NV-aware transcript $\mathbf{y}=(y_1,\ldots,y_M)$ containing words and inline NV markers.
Its training objective mirrors Eq.~(1):
\begin{equation}
\mathcal{L}_{\mathrm{tr}}=-\sum_{m=1}^{M}\log p(y_m\mid y_{<m},\boldsymbol{\mu}).
\end{equation}
This model is used only for evaluation, not in the TTS generation path.

The \emph{waveform path} renders the semantic tokens with the frozen flow decoder and vocoder, then applies Whisper-D\footnote{\small\href{https://huggingface.co/jordand/whisper-d-v1a}{huggingface.co/jordand/whisper-d-v1a}}~\cite{darefsky2024parakeet} to the waveform.
It therefore scores the complete synthesis pipeline.
The \emph{token path} applies our transcriber directly to the LM output, bypassing waveform rendering and its potential errors.
Both paths produce an NV-aware transcript.
We align each output with the reference prompt in text space and compute event-level precision, recall, and F1.
Agreement between the two paths indicates that an effect originates in semantic-token generation and persists after waveform rendering; a gap helps localize errors to the acoustic back end. Crucially, this study measures laughter event retrieval in TTS rather than perceptual audio quality. Evaluating laughter retrieval under this setup serves as a controlled proxy to validate whether cross-event knowledge transfer successfully translates into predictable NV generation.

\section{Experiments}

\subsection{Experimental Setup}
\textbf{Data and backbone.}
We use Libri-Light~\cite{kahn2020libri} as NV-neutral pre-training data and an internal emotional TTS corpus as the source of NV subsets.
The backbone is CosyVoice 2-0.5B~\cite{du2024cosyvoice}, and each run uses eight H100 GPUs.
To reduce native NV priors, we reinitialize the autoregressive TTS LM and pre-train it on approximately 27k hours of neutral Libri-Light speech.
Following CosyVoice 2, we initialize the backbone from Qwen2-0.5B~\cite{qwen2}, randomly initialize the speech-token layers and NV embeddings, and freeze the FSQ tokenizer, flow decoder, and vocoder.
Pre-training uses AdamW with weight decay $0.01$, attention dropout $0.1$, gradient clipping at $1.0$, and a cosine learning-rate schedule from $5\times10^{-5}$ to $10^{-6}$ with 20k warm-up steps.
The epoch-20 checkpoint obtains 4.7\% WER on 50 speaker-disjoint LibriTTS test-clean utterances and produces no NVs on the evaluation prompts.

\textbf{Fine-tuning sets.}
We normalize transcripts to eight canonical NV tags; examples include laughter, breath, sigh, cough, and lip-smack.
Data budgets count tagged NV occurrences; the figures refer to each occurrence as one example.
Unambiguous variants map to a canonical tag; unsupported or ambiguous annotations are removed.
Pure-laughter utterances contain laughter and no other NV token.
Donor utterances contain the selected donor event and no other NV token.
The neutral control adds 3,000 Libri-Light utterances.
All pools exclude a deterministic 1\% speaker-disjoint development split.
We fine-tune for 10 epochs with AdamW, learning rate $10^{-5}$, weight decay $0.01$, $(\beta_1,\beta_2)=(0.9,0.95)$, $\epsilon=10^{-8}$, attention dropout $0.1$, and gradient clipping at $1.0$.
The learning rate decays to $10^{-6}$ after a 5\% warm-up, and the step budget scales with set size so that every example is seen for 10 epochs.

\textbf{Transcriber training.}
Our evaluator uses the released CosyVoice 2 LM with a Qwen2-0.5B backbone.
We train it on the internal emotional corpus and English Libri-Light for 100k steps using AdamW, weight decay $0.01$, attention dropout $0.1$, and a learning rate that warms up for 5k steps to $10^{-4}$ and decays to $10^{-5}$.
The checkpoint reaches 4.88\% WER on LibriTTS test-clean.

\textbf{Evaluation set.}
Each checkpoint is evaluated on the same 400 prompts, including 100 laughter-positive prompts.
We score generated waveforms with Whisper-D and LM outputs with the semantic-token transcriber.
On oracle audio and oracle semantic tokens from this set, the two evaluators obtain laughter-retrieval F1 scores of 0.712 and 0.813, respectively.
Inference uses deterministic FP32 computation.

\subsection{Results}
\subsubsection{Target-Only Scaling}
\label{sec:scaling_exps}
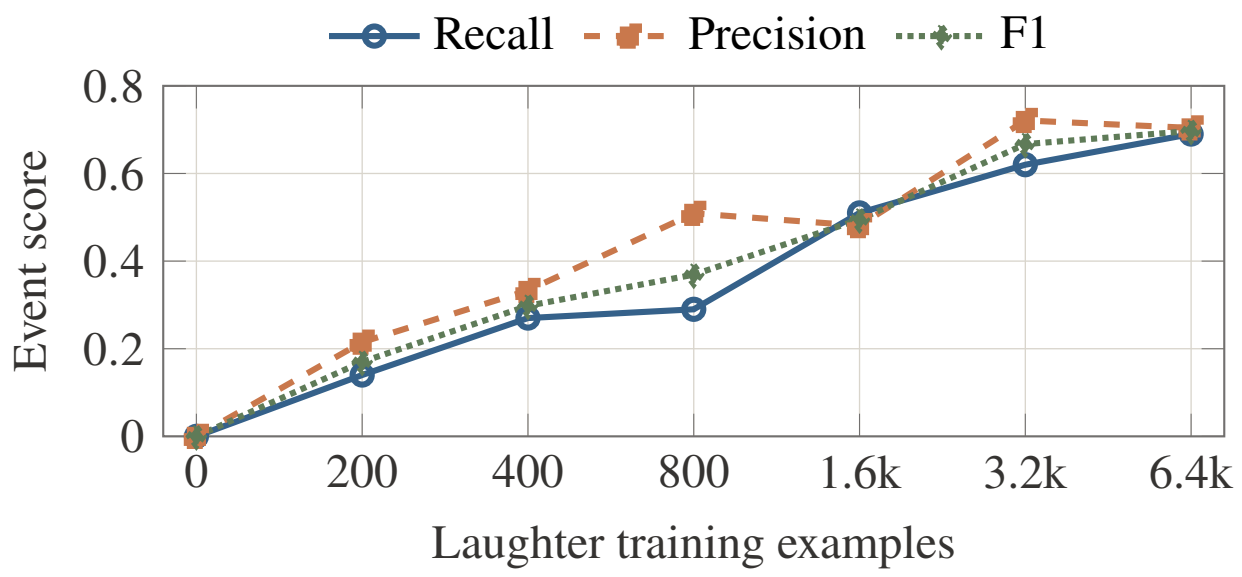
\begin{figure}[t]
  \centering
  \begin{tikzpicture}
\begin{axis}[
  nvaxis,
  width=\columnwidth,
  height=3.9cm,
  xmin=-0.2,xmax=6.2,
  ymin=0,ymax=0.8,
  xtick={0,1,2,3,4,5,6},
  xticklabels={0,200,400,800,1.6k,3.2k,6.4k},
  ytick={0,0.2,0.4,0.6,0.8},
  xlabel={Laughter training examples},
  ylabel={Event score},
  legend columns=3,
  legend style={at={(0.5,1.02)},anchor=south},
]
\addplot[color=nvblue,mark=o] coordinates {(0,0) (1,.140) (2,.270) (3,.290) (4,.510) (5,.620) (6,.690)};
\addlegendentry{Recall}
\addplot[color=nvorange,mark=square*,dashed] coordinates {(0,0) (1,.215) (2,.333) (3,.509) (4,.481) (5,.721) (6,.704)};
\addlegendentry{Precision}
\addplot[color=nvgreen,mark=diamond*,densely dotted] coordinates {(0,0) (1,.170) (2,.298) (3,.369) (4,.495) (5,.667) (6,.697)};
\addlegendentry{F1}
\end{axis}
\end{tikzpicture}
  \caption{Waveform-level laughter scaling. Performance rises quickly through 3,200 target examples and changes little from 3,200 to 6,400.}
  \label{fig:scaling}
\end{figure}

Figure~\ref{fig:scaling} establishes the laughter-only baseline.
F1 rises from 0.170 at 200 examples to 0.495 at 1,600 and 0.667 at 3,200.
Doubling the data once more yields only 0.030 absolute improvement, reaching 0.697 at 6,400.
The curve places 200 examples in a data-starved regime and 1,600 near saturation.

\subsubsection{Cross-Token Transfer}
\begin{table}[t]
  \centering
  \caption{Cross-token transfer at the 200-laughter anchor. Each donor contributes 3,200 NV examples.}
  \label{tab:transfer_anchor200}
  \small
  \renewcommand{\arraystretch}{1.05}
  \begin{tabular*}{\columnwidth}{@{\extracolsep{\fill}}l ccc ccc@{}}
    \toprule
    & \multicolumn{3}{c}{\textbf{\textcolor{nvblue}{Waveform}}} & \multicolumn{3}{c}{\textbf{\textcolor{nvpurple}{Semantic token}}} \\
    \cmidrule(lr){2-4} \cmidrule(lr){5-7}
    \textbf{Donor} & \textbf{P} & \textbf{R} & \textbf{F1} & \textbf{P} & \textbf{R} & \textbf{F1} \\
    \midrule
    None      & .215 & .140 & .170 & .333 & .241 & .280 \\
    Breath    & \textbf{.721} & .310 & .434 & .746 & .312 & .440 \\
    Sigh      & .628 & .270 & .378 & .831 & .348 & .490 \\
    \rowcolor{nvorange!8}
    Cough     & .622 & \textbf{.460} & \textbf{.529} & .731 & \textbf{.482} & \textbf{.581} \\
    Lip-smack & .630 & .290 & .397 & \textbf{.857} & .340 & .487 \\
    \addlinespace[1pt]
    \rowcolor{black!3}
    Neutral   & .083 & .020 & .032 & .250 & .050 & .083 \\
    \bottomrule
  \end{tabular*}
\end{table}

\begin{table}[t]
  \centering
  \caption{Cross-token transfer at the 1,600-laughter anchor. Each donor contributes 3,200 NV examples.}
  \label{tab:transfer_anchor1600}
  \small
  \renewcommand{\arraystretch}{1.05}
  \begin{tabular*}{\columnwidth}{@{\extracolsep{\fill}}l ccc ccc@{}}
    \toprule
    & \multicolumn{3}{c}{\textbf{\textcolor{nvblue}{Waveform}}} & \multicolumn{3}{c}{\textbf{\textcolor{nvpurple}{Semantic token}}} \\
    \cmidrule(lr){2-4} \cmidrule(lr){5-7}
    \textbf{Donor} & \textbf{P} & \textbf{R} & \textbf{F1} & \textbf{P} & \textbf{R} & \textbf{F1} \\
    \midrule
    None      & .481 & .510 & .495 & .678 & .582 & .626 \\
    Breath    & \textbf{.759} & .630 & .689 & .859 & .560 & .678 \\
    Sigh      & .707 & .580 & .637 & .865 & .589 & .700 \\
    \rowcolor{nvorange!8}
    Cough     & .733 & \textbf{.660} & \textbf{.695} & .894 & .596 & .715 \\
    \rowcolor{nvpurple!6}
    Lip-smack & .747 & .650 & \textbf{.695} & \textbf{.914} & \textbf{.603} & \textbf{.726} \\
    \addlinespace[1pt]
    \rowcolor{black!3}
    Neutral   & .538 & .500 & .518 & .741 & .567 & .643 \\
    \bottomrule
  \end{tabular*}
\end{table}

Tables~\ref{tab:transfer_anchor200} and~\ref{tab:transfer_anchor1600} compare donor tokens at the two laughter anchors.
With 200 laughter examples, every NV donor improves on the 0.170 waveform F1 baseline, while neutral speech reduces F1 to 0.032.
Cough performs best at 0.529, a $3.1\times$ relative improvement that also exceeds the 0.495 achieved with 1,600 laughter examples alone.
At the 1,600-example anchor, cough and lip-smack each reach 0.695, nearly matching the 0.697 result from 6,400 target examples.
The neutral control confirms that arbitrary speech is insufficient. 

The largest change appears in precision.
At the 200-example anchor, cough raises waveform precision from 0.215 to 0.622 and recall from 0.140 to 0.460.
The token evaluator shows the same pattern: precision rises from 0.333 to 0.731, while recall rises from 0.241 to 0.482.
Donor NVs therefore help the semantic LM suppress spurious laughter while recovering more requested events.

\begin{figure}[t]
  \centering
  \begin{tikzpicture}
\begin{groupplot}[
  group style={group size=2 by 1,horizontal sep=0.75cm},
  nvaxis,
  width=0.53\columnwidth,
  height=3.8cm,
  xmin=-0.2,xmax=5.2,
  ymin=0,ymax=0.8,
  xtick={0,2,5},
  xticklabels={0,400,3.2k},
  ytick={0,0.2,0.4,0.6,0.8},
  xlabel={Cough donor examples},
]
\nextgroupplot[
  title={Waveform},
  ylabel={Event score},
  legend columns=3,
  legend style={at={(1.12,1.18)},anchor=south},
]
\addplot[color=nvblue,mark=o] coordinates {(0,.140) (1,.190) (2,.190) (3,.280) (4,.320) (5,.460)};
\addlegendentry{R}
\addplot[color=nvorange,mark=square*,dashed] coordinates {(0,.215) (1,.345) (2,.302) (3,.583) (4,.508) (5,.622)};
\addlegendentry{P}
\addplot[color=nvgreen,mark=diamond*,densely dotted] coordinates {(0,.170) (1,.245) (2,.233) (3,.378) (4,.393) (5,.529)};
\addlegendentry{F1}
\nextgroupplot[title={Semantic token},yticklabels={}]
\addplot[color=nvblue,mark=o] coordinates {(0,.241) (1,.255) (2,.177) (3,.284) (4,.340) (5,.461)};
\addplot[color=nvorange,mark=square*,dashed] coordinates {(0,.333) (1,.439) (2,.379) (3,.556) (4,.676) (5,.739)};
\addplot[color=nvgreen,mark=diamond*,densely dotted] coordinates {(0,.280) (1,.323) (2,.242) (3,.376) (4,.453) (5,.568)};
\end{groupplot}
\end{tikzpicture}
  \caption{Cough-donor scaling with 200 laughter examples. Small multiples separate the waveform and semantic-token evaluators so precision, recall, and F1 remain legible.}
  \label{fig:cough_scaling}
\end{figure}
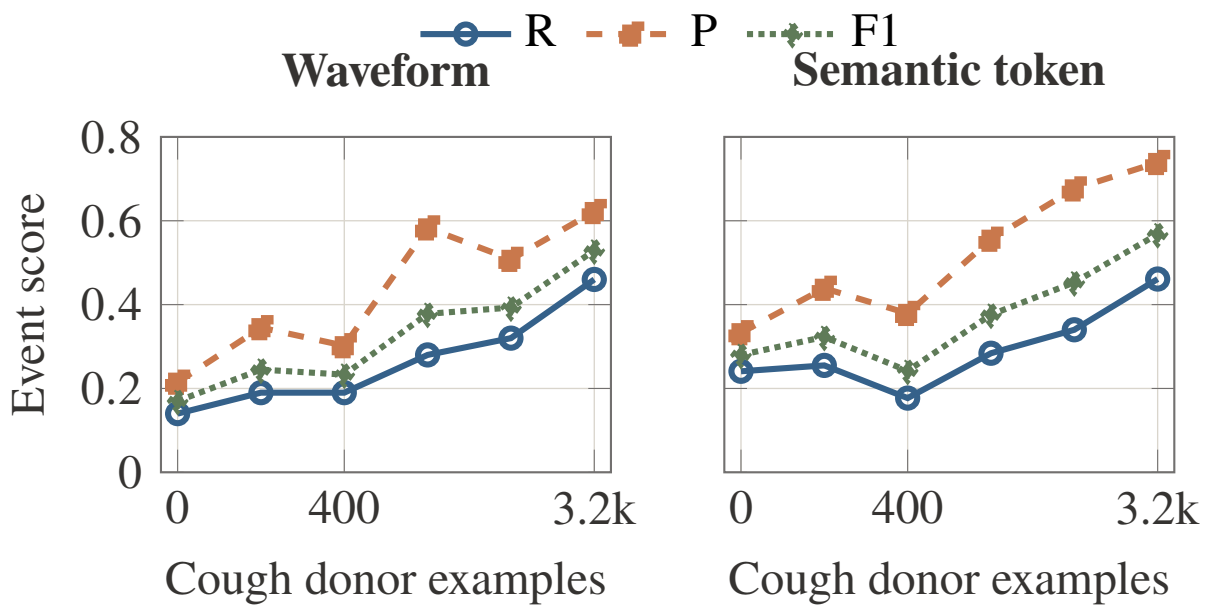

Figure~\ref{fig:cough_scaling} varies the cough donor budget.
The overall trend is positive but not strictly monotonic: waveform F1 changes from 0.245 at 200 cough examples to 0.233 at 400, then reaches 0.529 at 3,200.
The token path follows the same broad pattern and reaches 0.568, although it generally gives a higher absolute score.
The agreement localizes the effect to semantic-token generation and leaves a decoder gap.

\begin{figure}[t]
  \centering
  \includegraphics[width=\columnwidth]{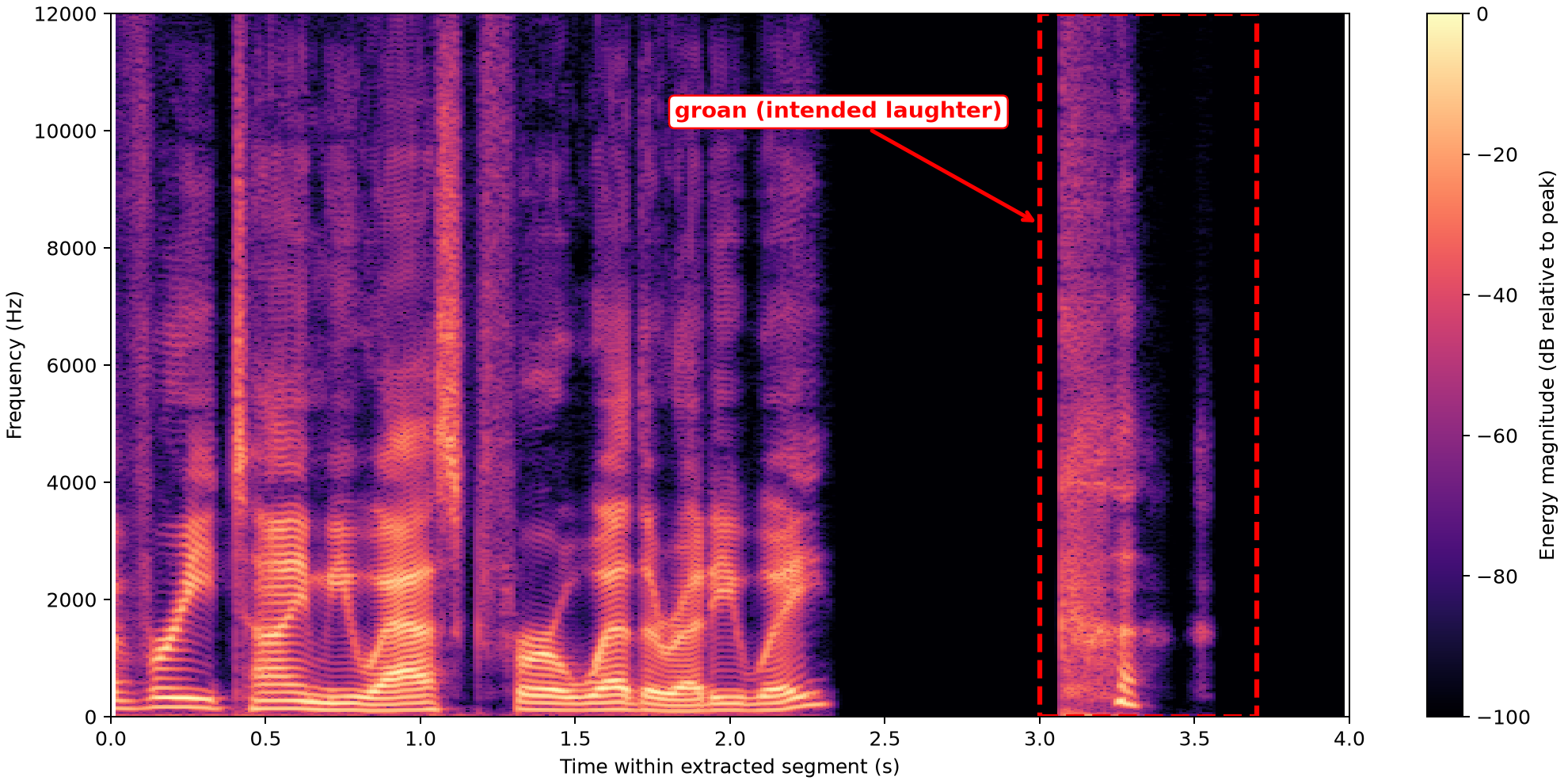}
  \vspace{-1mm}
  \centerline{\small (a) 1,600 laughter examples}
  \vspace{1mm}
  \includegraphics[width=\columnwidth]{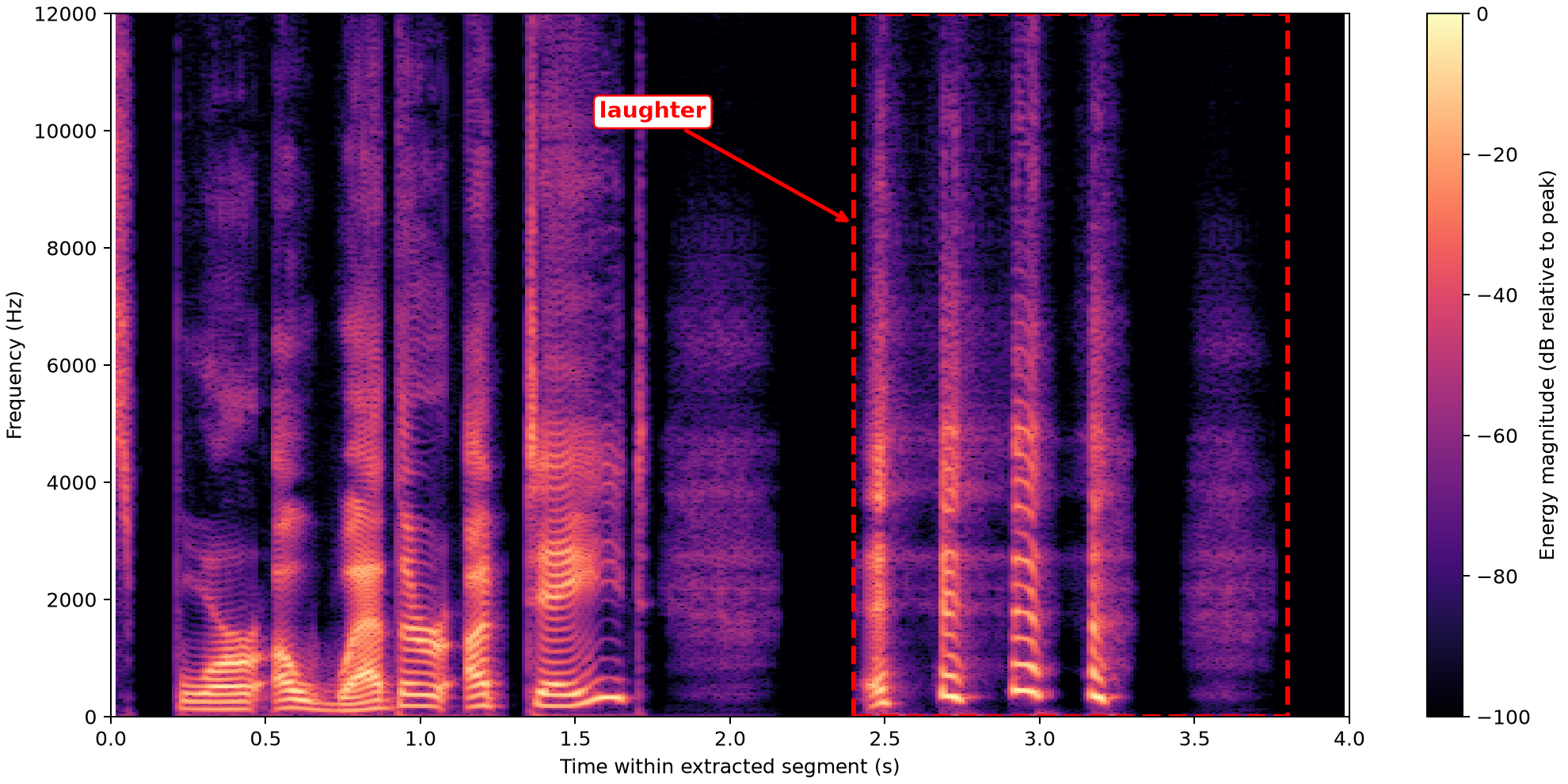}
  \vspace{-1mm}
  \centerline{\small (b) 1,600 laughter + 3,200 cough examples}
  \caption{Spectrograms for the same prompt. The red boxes mark the requested laughter interval. The target-only model produces diffuse aspiration, whereas the cough-augmented model produces structured bursts between 2.4 and 3.8 s.}
  \vspace{-15pt}
  \label{fig:spectrogram_vertical}
\end{figure}

The spectrograms in Fig.~\ref{fig:spectrogram_vertical} illustrate the same difference.
For the prompt ``\ldots ask for a weather update? \texttt{[laughter]} \ldots'', the target-only model produces diffuse, groan-like aspiration in the labeled interval.
Adding cough data yields a sequence of clearer bursts.
This pattern suggests shared respiratory and glottal dynamics, pending listening tests.

\vspace{-10pt}
\section{Conclusion}
We studied cross-token transfer while holding the TTS architecture fixed.
Laughter-only performance showed diminishing returns beyond 3,200 examples.
With 200 laughter examples, adding 3,200 cough examples increased waveform F1 from 0.170 to 0.529, exceeding 1,600 laughter examples alone.
Waveform and semantic-token evaluation agreed on the main trends, placing the gain before waveform rendering and showing that it largely survives the acoustic decoder.
We study one target and one backbone and measure event retrieval rather than perceptual quality; broader NV pairs and listening tests remain important.

\bibliographystyle{IEEEbib}
\bibliography{refs}

@inproceedings{wu2024laugh,
  title={Laugh now cry later: Controlling time-varying emotional states of flow-matching-based zero-shot text-to-speech},
  author={Wu, Haibin and Wang, Xiaofei and Eskimez, Sefik Emre and others},
  booktitle={2024 IEEE Spoken Language Technology Workshop (SLT)},
  pages={690--697},
  year={2024},
  organization={IEEE}
}

@book{provine2000laughter,
  title={Laughter: A scientific investigation},
  author={Provine, Robert R},
  year={2000},
  publisher={Penguin}
}

@article{scherer2003vocal,
  title={Vocal communication of emotion: A review of research paradigms},
  author={Scherer, Klaus R},
  journal={Speech communication},
  volume={40},
  number={1-2},
  pages={227--256},
  year={2003},
  publisher={Elsevier}
}

@inproceedings{trouvain2014laughing,
  title={Laughing, breathing, clicking-The prosody of nonverbal vocalizations},
  author={Trouvain, J{\"u}rgen},
  booktitle={Proceedings of the 7th international conference on Speech Prosody},
  pages={598--602},
  year={2014}
}

@misc{darefsky2024parakeet,
    author = {Darefsky, Jordan and Zhu, Ge and Duan, Zhiyao},
    title = {Parakeet},
    year = {2024},
    url = {https://jordandarefsky.com/blog/2024/parakeet/}
}

@article{liao2025nvspeech,
  title={NVSpeech: An Integrated and Scalable Pipeline for Human-Like Speech Modeling with Paralinguistic Vocalizations},
  author={Liao, Huan and Ni, Qinke and Wang, Yuancheng and Lu, Yiheng and Zhan, Haoyue and Xie, Pengyuan and Zhang, Qiang and Wu, Zhizheng},
  journal={arXiv preprint arXiv:2508.04195},
  year={2025}
}

@article{ye2025nonverbalspeech,
  title={A Scalable Pipeline for Enabling Non-Verbal Speech Generation and Understanding},
  author={Ye, Runchuan and Zhou, Yixuan and Yu, Renjie and Lin, Zijian and Li, Kehan and Li, Xiang and Liu, Xin and Zeng, Guoyang and Wu, Zhiyong},
  journal={arXiv preprint arXiv:2508.05385},
  year={2025}
}

@article{bai2025synparaspeech,
  title={SynParaSpeech: Automated Synthesis of Paralinguistic Datasets for Speech Generation and Understanding},
  author={Bai, Bingsong and Lu, Qihang and Yang, Wenbing and Sun, Zihan and Hou, Yueran and Jia, Peilei and Pu, Songbai and Fu, Ruibo and Gao, Yingming and Li, Ya and Gao, Jun},
  journal={arXiv preprint arXiv:2509.14946},
  year={2025}
}

@article{borisov2025nonverbaltts,
  title={NonverbalTTS: A Public English Corpus of Text-Aligned Nonverbal Vocalizations with Emotion Annotations for Text-to-Speech},
  author={Borisov, Maksim and Spirin, Egor and Diatlova, Daria},
  journal={arXiv preprint arXiv:2507.13155},
  year={2025}
}

@inproceedings{kahn2020libri,
  title={Libri-light: A benchmark for asr with limited or no supervision},
  author={Kahn, Jacob and Riviere, Morgane and Zheng, Weiyi and others},
  booktitle={ICASSP 2020-2020 IEEE international conference on acoustics, speech and signal processing (ICASSP)},
  pages={7669--7673},
  year={2020},
  organization={IEEE}
}

@article{du2024cosyvoice,
  title={CosyVoice 2: Scalable streaming speech synthesis with large language models},
  author={Du, Zhihao and Wang, Yuxuan and Chen, Qian and others},
  journal={arXiv preprint arXiv:2412.10117},
  year={2024}
}

@article{qwen2,
      title={Qwen2 Technical Report}, 
      author={Yang, An and Yang, Baosong and Hui, Binyuan and others},
      journal={arXiv preprint arXiv:2407.10671},
      year={2024}
}

@inproceedings{wu2025smiipnv,
  title={{SMIIP-NV}: A Multi-Annotation Non-Verbal Expressive Speech Corpus in {Mandarin} for LLM-Based Speech Synthesis},
  author={Wu, Zhuojun and Liu, Dong and Liu, Juan and others},
  booktitle={Proceedings of the 33rd ACM International Conference on Multimedia},
  pages={12564--12570},
  year={2025}
}

@inproceedings{mai2026mnv17,
  title={{MNV-17}: A High-Quality Performative {Mandarin} Dataset for Nonverbal Vocalization Recognition in Speech},
  author={Mai, Jialong and Ji, Jinxin and Xing, Xiaofen and others},
  booktitle={IEEE International Conference on Acoustics, Speech and Signal Processing (ICASSP)},
  pages={18312--18316},
  year={2026}
}

@article{yang2026wesr,
  title={{WESR}: Scaling and Evaluating Word-Level Event-Speech Recognition},
  author={Yang, Chenchen and Huang, Kexin and Fan, Liwei and Tu, Qian and Jiang, Botian and Zhang, Dong and Yin, Linqi and Li, Shimin and Fei, Zhaoye and Cheng, Qinyuan and Qiu, Xipeng},
  journal={arXiv preprint arXiv:2601.04508},
  year={2026}
}

@article{yang2026beyond,
  title={Beyond Words: Towards Effective Modeling of Non-Verbal Vocalizations in ASR},
  author={Yang, Gene and Wu, Haibin and Su, Peng and Huang, Ruizhe and Shon, Suwon and Do, Bach and Niu, Minxue and Ni, Zhaoheng and Li, Shang-Wen and Metze, Florian and Adi, Yossi and Sun, Ming and Liu, Yuzong},
  journal={arXiv preprint arXiv:2607.01563},
  year={2026}
}

@article{ni2026nvbench,
  title={{NV-Bench}: Benchmark of Nonverbal Vocalization Synthesis for Expressive Text-to-Speech Generation},
  author={Ni, Qinke and Liao, Huan and Chen, Dekun and Wang, Yuxiang and Wu, Zhizheng},
  journal={arXiv preprint arXiv:2603.15352},
  year={2026}
}

@article{xue2026nvvsuperbench,
  title={{NVV-SuperBench}: Beyond Words, Beyond Quality---Benchmarking Nonverbal Vocalizations in Speech Generation},
  author={Xue, Liumeng and Bian, Weizhen and Pan, Jiahao and Wu, Wenxuan and Ren, Yilin and Kang, Boyi and Hu, Jingbin and Ma, Ziyang and Wang, Shuai and Qian, Xinyuan and Lee, Hung-yi and Guo, Yike},
  journal={arXiv preprint arXiv:2604.16211},
  year={2026}
}

@article{li2026preference,
  title={Preference Optimization for Non-Verbal Vocalization Synthesis},
  author={Li, Haoyang and Xu, Chenglin and Zhao, Junchuan and Cao, Yuang and Xue, Liumeng and Guo, Yiwen and Chng, Eng Siong},
  journal={arXiv preprint arXiv:2608.24163},
  year={2026}
}

@inproceedings{inaguma2018joint,
  author    = {Hirofumi Inaguma and Masato Mimura and Koji Inoue and Kazuyoshi Yoshii and Tatsuya Kawahara},
  title     = {An End-to-End Approach to Joint Social Signal Detection and Automatic Speech Recognition},
  booktitle = {Proc. IEEE ICASSP},
  pages     = {5199--5203},
  year      = {2018}
}

@article{shione2023automatic,
  author  = {Nagisa Shione and Tatsuya Kawahara},
  title   = {Automatic Speech Recognition Using Linguistic and Verbal/Nonverbal Information},
  journal = {IEEE Access},
  volume  = {11},
  pages   = {138985--138995},
  year    = {2023}
}
\end{document}